\documentclass[11pt]{article}
\usepackage{coling08}
\usepackage{times}
\usepackage{latexsym}
\newenvironment{myitemize}{   
\vspace{-0.3\baselineskip}
\begin{itemize}
  \setlength{\topsep}{0pt}
  \setlength{\itemsep}{1pt}
  \setlength{\parskip}{0pt}
  \setlength{\parsep}{0pt}
  \setlength{\partopsep}{0pt}
}{
\end{itemize}
\vspace{-0.2\baselineskip}}

\nocopyrightfn
\title{A Uniform Approach to Analogies, Synonyms, Antonyms, \\
and Associations}

\author{Peter D. Turney\\
  National Research Council of Canada\\
  Institute for Information Technology\\
  M50 Montreal Road\\
  Ottawa, Ontario, Canada\\
  K1A 0R6\\
  {\tt peter.turney@nrc-cnrc.gc.ca}}

\date{}

\begin{document}
\maketitle
\begin{abstract}
Recognizing analogies, synonyms, anto\-nyms, and associations appear to be four
distinct tasks, requiring distinct NLP algorithms. In the past, the four
tasks have been treated independently, using a wide variety of algorithms.
These four semantic classes, however, are a tiny sample of the full
range of semantic phenomena, and we cannot afford to create ad hoc algorithms
for each semantic phenomenon; we need to seek a unified approach.
We propose to subsume a broad range of phenomena under analogies.
To limit the scope of this paper, we restrict our attention to the subsumption
of synonyms, antonyms, and associations. We introduce a supervised corpus-based
machine learning algorithm for classifying analogous word pairs, and we
show that it can solve multiple-choice SAT analogy questions, TOEFL
synonym questions, ESL synonym-antonym questions, and similar-associated-both
questions from cognitive psychology.
\end{abstract}

\section{Introduction}

A pair of words (petrify:stone) is \emph{analogous} to another pair (vaporize:gas)
when the semantic relations between the words in the first pair are highly
similar to the relations in the second pair. Two words (levied and imposed) 
are \emph{synonymous} in a context (levied a tax) when they can be interchanged
(imposed a tax), they are are \emph{antonymous} when they have opposite meanings
(black and white), and they are \emph{associated} when they tend to co-occur
(doctor and hospital).

On the surface, it appears that these are four distinct semantic classes,
requiring distinct NLP algorithms, but we propose a uniform approach to all
four. We subsume synonyms, antonyms, and associations under analogies.
In essence, we say that $X$ and $Y$ are antonyms when the pair $X$:$Y$ is
analogous to the pair black:white, $X$ and $Y$ are synonyms when they are
analogous to the pair levied:imposed, and $X$ and $Y$ are associated when
they are analogous to the pair doctor:hospital.

There is past work on recognizing analogies \cite{reitman65},
synonyms \cite{landauer97}, antonyms \cite{lin03}, and associations
\cite{lesk69}, but each of these four tasks has been examined
separately, in isolation from the others. As far as we know, the
algorithm proposed here is the first attempt to deal with all four
tasks using a uniform approach. We believe that it is important
to seek NLP algorithms that can handle a broad range of semantic
phenomena, because developing a specialized algorithm for each
phenomenon is a very inefficient research strategy.

It might seem that a lexicon, such as WordNet \cite{fellbaum98},
contains all the information we need to handle these four tasks.
However, we prefer to take a corpus-based approach to semantics.
Veale \shortcite{veale04} used WordNet to answer 374 multiple-choice
SAT analogy questions, achieving an accuracy of 43\%, but the
best corpus-based approach attains an accuracy of 56\%
\cite{turney06}. Another reason to prefer a corpus-based approach
to a lexicon-based approach is that the former requires less human
labour, and thus it is easier to extend to other languages.

In Section~\ref{sec:analogy-perception}, we describe our algorithm
for recognizing analogies. We use a standard supervised machine
learning approach, with feature vectors based on the frequencies
of patterns in a large corpus. We use a support vector machine (SVM)
to learn how to classify the feature vectors \cite{platt98,witten99}.

Section~\ref{sec:experiments} presents four sets of experiments.
We apply our algorithm for recognizing analogies to multiple-choice
analogy questions from the SAT college entrance test, multiple-choice
synonym questions from the TOEFL (test of English as a foreign language),
ESL (English as a second language) practice questions for
distinguishing synonyms and antonyms, and a set of word pairs
that are labeled \emph{similar}, \emph{associated}, and \emph{both},
developed for experiments in cognitive psychology.

We discuss the results of the experiments in Section~\ref{sec:discussion}.
The accuracy of the algorithm is competitive with other systems, but the
strength of the algorithm is that it is able to handle all four tasks, with
no tuning of the learning parameters to the particular task. It performs well,
although it is competing against specialized algorithms, developed
for single tasks.

Related work is examined in Section~\ref{sec:related} and limitations
and future work are considered in Section~\ref{sec:limitations}. We
conclude in Section~\ref{sec:conclusion}.

\section{Classifying Analogous Word Pairs}
\label{sec:analogy-perception}

An analogy, $A$:$B$::$C$:$D$, asserts that $A$ is to $B$
as $C$ is to $D$; for example, traffic:street::water:riverbed
asserts that traffic is to street as water is to riverbed;
that is, the semantic relations between traffic and street are
highly similar to the semantic relations between water and
riverbed. We may view the task of recognizing word analogies
as a problem of classifying word pairs (see Table~\ref{tab:pair-classes}).

\begin{table}[htbp]
\centering
\begin{tabular*}{0.75\linewidth}{@{\extracolsep{\fill}}ll}
\hline
Word pair & Class label \\
\hline
carpenter:wood     & artisan:material \\
mason:stone        & artisan:material \\
potter:clay        & artisan:material \\
glassblower:glass  & artisan:material \\
traffic:street     & entity:carrier \\
water:riverbed     & entity:carrier \\
packets:network    & entity:carrier \\
gossip:grapevine   & entity:carrier \\
\hline
\end{tabular*}
\caption {Examples of how the task of recognizing word analogies may be
viewed as a problem of classifying word pairs.}
\label{tab:pair-classes}
\end{table}

We approach this as a standard classification problem for supervised
machine learning. The algorithm takes as input a training set of word
pairs with class labels and a testing set of word pairs without labels.
Each word pair is represented as a vector in a feature space and a supervised
learning algorithm is used to classify the feature vectors. The elements in
the feature vectors are based on the frequencies of automatically defined
patterns in a large corpus. The output of the algorithm is an assignment
of labels to the word pairs in the testing set. For some of the experiments,
we select a unique label for each word pair; for other experiments, we
assign probabilities to each possible label for each word pair.

For a given word pair, such as mason:stone, the first step is to
generate morphological variations, such as masons:stones. In the
following experiments, we use \emph{morpha} (morphological analyzer) and
\emph{morphg} (morphological generator) for morphological processing
\cite{minnen01}.\footnote{http://www.informatics.susx.ac.uk/research/groups/nlp/
carroll/morph.html.}

The second step is to search in a large corpus for all phrases of
the following form:

\vspace{1pt}
\noindent ``[0 to 1 words] $X$ [0 to 3 words] $Y$ [0 to 1 words]''

\noindent In this template, $X$:$Y$ consists of morphological variations
of the given word pair, in either order; for example, mason:stone, stone:mason,
masons:stones, and so on. A typical phrase for mason:stone would
be ``the mason cut the stone with''. We then normalize all of the
phrases that are found, by using \emph{morpha} to remove suffixes.

The template we use here is similar to Turney \shortcite{turney06},
but we have added extra context words before the $X$ and after the $Y$.
Our morphological processing also differs from Turney \shortcite{turney06}.
In the following experiments, we search in a corpus of $5 \times 10^{10}$
words (about 280 GB of plain text), consisting of web pages gathered
by a web crawler.\footnote{The corpus was collected by Charles Clarke,
University of Waterloo. We can provide copies on request.} To retrieve
phrases from the corpus, we use Wumpus \cite{buettcher05}, an efficient
search engine for passage retrieval
from large corpora.\footnote{http://www.wumpus-search.org/.}

The next step is to generate patterns from all of the phrases that
were found for all of the input word pairs (from both the training
and testing sets). To generate patterns from a phrase, we replace the given
word pairs with variables, $X$ and $Y$, and we replace the remaining words
with a wild card symbol (an asterisk) or leave them as they are.
For example, the phrase ``the mason cut the stone with'' yields the
patterns ``the $X$ cut * $Y$ with'', ``* $X$ * the $Y$ *'', and so on.
If a phrase contains $n$ words, then it yields $2^{(n-2)}$ patterns.

Each pattern corresponds to a feature in the feature vectors that we
will generate. Since a typical input set of word pairs yields
millions of patterns, we need to use feature selection, to reduce
the number of patterns to a manageable quantity. For each pattern,
we count the number of input word pairs that generated the pattern.
For example, ``* $X$ cut * $Y$ *'' is generated by both mason:stone
and carpenter:wood. We then sort the patterns in descending order
of the number of word pairs that generated them. If there are $N$
input word pairs (and thus $N$ feature vectors, including both the
training and testing sets), then we select
the top $kN$ patterns and drop the remainder. In the
following experiments, $k$ is set to 20. The algorithm is not
sensitive to the precise value of $k$.

The reasoning behind the feature selection algorithm is that shared
patterns make more useful features than rare patterns.
The number of features ($kN$) depends on the number of word pairs ($N$),
because, if we have more feature vectors, then we need more features to
distinguish them. Turney \shortcite{turney06} also selects patterns
based on the number of pairs that generate them, but the number of
selected patterns is a constant (8000), independent of the number of
input word pairs.

The next step is to generate feature vectors, one vector for each
input word pair. Each of the $N$ feature vectors has $kN$ elements,
one element for each selected pattern. The value of an element
in a vector is given by the logarithm of the frequency in the corpus
of the corresponding pattern for the given word pair. For example,
suppose the given pair is mason:stone and the pattern is
``* $X$ cut * $Y$ *''. We look at the normalized phrases that
we collected for mason:stone and we count how many match this
pattern. If $f$ phrases match the pattern, then the value of
this element in the feature vector is $\log(f+1)$ (we add $1$
because $\log(0)$ is undefined). Each feature vector is then
normalized to unit length. The normalization ensures that
features in vectors for high-frequency word pairs (traffic:street) are
comparable to features in vectors for low-frequency word pairs (water:riverbed).

Now that we have a feature vector for each input word pair,
we can apply a standard supervised learning algorithm.
In the following experiments, we use a sequential minimal optimization
(SMO) support vector machine (SVM) with a radial basis function (RBF)
kernel \cite{platt98}, as implemented in Weka (Waikato Environment for
Knowledge Analysis) \cite{witten99}.\footnote{http://www.cs.waikato.ac.nz/ml/weka/.}
The algorithm generates probability estimates for each class by fitting logistic
regression models to the outputs of the SVM. We disable the normalization option
in Weka, since the vectors are already normalized to unit length.
We chose the SMO RBF algorithm because it is fast, robust, and
it easily handles large numbers of features.

For convenience, we will refer to the above algorithm as PairClass. In
the following experiments, PairClass is applied to each of the four
problems with no adjustments or tuning to the specific problems. Some work
is required to fit each problem into the general framework of PairClass
(supervised classification of word pairs) but the core algorithm is the
same in each case.

\section{Experiments}
\label{sec:experiments}

This section presents four sets of experiments, with analogies,
synonyms, antonyms, and associations. We explain how each task
is treated as a problem of classifying analogous word pairs,
we give the experimental results, and we discuss past work
on each of the four tasks.

\subsection{SAT Analogies}

In this section, we apply PairClass to the task of recognizing
analogies. To evaluate the performance, we use a set of 374 multiple-choice
questions from the SAT college entrance exam. Table~\ref{tab:sat} shows
a typical question. The target pair is called the \emph{stem}.
The task is to select the choice pair that is most analogous to
the stem pair.

\begin{table}[htbp]
\centering
\begin{tabular*}{0.75\linewidth}{@{\extracolsep{\fill}}lll}
\hline
Stem:      &       & mason:stone \\
\hline
Choices:   & (a)   & teacher:chalk \\
           & (b)   & carpenter:wood \\
           & (c)   & soldier:gun \\
           & (d)   & photograph:camera \\
           & (e)   & book:word \\
\hline
 Solution: & (b)   & carpenter:wood \\
\hline
\end{tabular*}
\caption {An example of a question from the 374 SAT analogy questions.}
\label{tab:sat}
\end{table}

The problem of recognizing
word analogies was first attempted with a system called Argus \cite{reitman65},
using a small hand-built semantic network with a spreading activation
algorithm. Turney et al. \shortcite{turneyetal03} used a combination of 13
independent modules. Veale \shortcite{veale04} used a spreading activation
algorithm with WordNet (in effect, treating WordNet as a semantic network).
Turney \shortcite{turney06} used a corpus-based algorithm.

We may view Table~\ref{tab:sat} as a binary classification problem, in which
mason:stone and carpenter:wood are positive examples and the remaining word
pairs are negative examples. The difficulty is that the labels of the
choice pairs must be hidden from the learning algorithm. That is, the
training set consists of one positive example (the stem pair) and the
testing set consists of five unlabeled examples (the five choice pairs).
To make this task more tractable, we randomly choose a stem pair from
one of the 373 other SAT analogy questions, and we assume that this
new stem pair is a negative example, as shown in Table~\ref{tab:sat-frame}.

\begin{table}[htbp]
\centering
\begin{tabular*}{\linewidth}{@{\extracolsep{\fill}}lll}
\hline
Word pair & Train or test & Class label \\
\hline
mason:stone        & train               & positive \\
tutor:pupil        & train               & negative \\
\hline
teacher:chalk      & test                & hidden \\
carpenter:wood     & test                & hidden \\
soldier:gun        & test                & hidden \\
photograph:camera  & test                & hidden \\
book:word          & test                & hidden \\
\hline
\end{tabular*}
\caption {How to fit a SAT analogy question into the framework of
supervised pair classification.}
\label{tab:sat-frame}
\end{table}

To answer the SAT question, we use PairClass to estimate
the probability that each testing example is positive, and we guess
the testing example with the highest probability.
Learning from a training set with only one positive example and one
negative example is difficult, since the learned model can be highly unstable.
To increase the stability, we repeat the learning process 10 times,
using a different randomly chosen negative training example each time.
For each testing word pair, the 10 probability estimates are averaged together.
This is a form of bagging \cite{breiman96}.

PairClass attains an accuracy of 52.1\%. For comparison, the
ACL Wiki lists 12 previously published results with the 374 SAT
analogy questions.\footnote{For more information, see
\emph{SAT Analogy Questions (State of the art)}
at http://aclweb.org/aclwiki/.} Only 2 of the 12 algorithms have
higher accuracy. The best previous result is an accuracy of 56.1\%
\cite{turney06}. Random guessing would yield an accuracy of 20\%.
The average senior high school student achieves 57\% correct
\cite{turney06}.

\subsection{TOEFL Synonyms}

Now we apply PairClass to the task of recognizing synonyms,
using a set of 80 multiple-choice synonym questions from the
TOEFL (test of English as a foreign language). A sample question
is shown in Table~\ref{tab:toefl}. The task is to select the choice
word that is most similar in meaning to the stem word.

\begin{table}[htbp]
\centering
\begin{tabular*}{0.6\linewidth}{@{\extracolsep{\fill}}lll}
\hline
Stem:      &       & levied \\
\hline
Choices:   & (a)   & imposed \\
           & (b)   & believed \\
           & (c)   & requested \\
           & (d)   & correlated \\
\hline
 Solution: & (a)   & imposed \\
\hline
\end{tabular*}
\caption {An example of a question from the 80 TOEFL questions.}
\label{tab:toefl}
\end{table}

Synonymy can be viewed
as a high degree of semantic similarity. The most common way to measure
semantic similarity is to measure the distance between words in WordNet
\cite{resnik95,jiang97,hirst98}. Corpus-based measures
of word similarity are also common \cite{lesk69,landauer97,turney01}.

We may view Table~\ref{tab:toefl} as a binary classification problem,
in which the pair levied:imposed is a positive example of the class
\emph{synonymous} and the other possible pairings are negative
examples, as shown in Table~\ref{tab:toefl-frame}.

\begin{table}[htbp]
\centering
\begin{tabular*}{0.65\linewidth}{@{\extracolsep{\fill}}lll}
\hline
Word pair & Class label \\
\hline
levied:imposed      & positive \\
levied:believed     & negative \\
levied:requested    & negative \\
levied:correlated   & negative \\
\hline
\end{tabular*}
\caption {How to fit a TOEFL question into the framework of
supervised pair classification.}
\label{tab:toefl-frame}
\end{table}

The 80 TOEFL questions yield 320 ($80 \times 4$) word pairs,
80 labeled positive and 240 labeled negative. We apply PairClass
to the word pairs using ten-fold cross-validation. In each
random fold, 90\% of the pairs are used for training and 10\% are
used for testing. For each fold, the model that is learned from the
training set is used to assign probabilities to the pairs in the
testing set. With ten separate folds, the ten non-overlapping
testing sets cover the whole dataset. Our guess for each TOEFL question
is the choice with the highest probability of being positive, when
paired with the corresponding stem.

PairClass attains an accuracy of 76.2\%. For comparison, the
ACL Wiki lists 15 previously published results with the 80 TOEFL
synonym questions.\footnote{For more information, see
\emph{TOEFL Synonym Questions (State of the art)}
at http://aclweb.org/aclwiki/.} Of the 15 algorithms, 8 have higher
accuracy and 7 have lower. The best previous result is an accuracy of
97.5\% \cite{turneyetal03}, obtained using a hybrid of four different
algorithms. Random guessing would yield an accuracy of 25\%. The average
foreign applicant to a US university achieves 64.5\% correct
\cite{landauer97}.

\subsection{Synonyms and Antonyms}

The task of classifying word pairs as either synonyms or antonyms
readily fits into the framework of supervised classification of word pairs.
Table~\ref{tab:antonyms} shows some examples from a set of
136 ESL (English as a second language) practice questions that
we collected from various ESL websites.

\begin{table}[htbp]
\centering
\begin{tabular*}{0.8\linewidth}{@{\extracolsep{\fill}}ll}
\hline
Word pair & Class label \\
\hline
galling:irksome             &  synonyms \\
yield:bend                  &  synonyms \\
naive:callow                &  synonyms \\
advise:suggest              &  synonyms \\
dissimilarity:resemblance   &  antonyms \\
commend:denounce            &  antonyms \\
expose:camouflage           &  antonyms \\
unveil:veil                 &  antonyms \\
\hline
\end{tabular*}
\caption {Examples of synonyms and antonyms from 136 ESL practice questions.}
\label{tab:antonyms}
\end{table}

Lin et al. \shortcite{lin03} distinguish synonyms from antonyms using
two patterns, ``from $X$ to $Y$'' and ``either $X$ or $Y$''. When $X$ and $Y$
are antonyms, they occasionally appear in a large corpus in one of these
two patterns, but it is very rare for synonyms to appear in these patterns.
Our approach is similar to Lin et al. \shortcite{lin03}, but we do not
rely on hand-coded patterns; instead, PairClass patterns are generated
automatically.

Using ten-fold cross-validation, PairClass attains an accuracy of
75.0\%. Always guessing the majority class would result in an accuracy
of 65.4\%. The average human score is unknown and there are no
previous results for comparison.

\subsection{Similar, Associated, and Both}

A common criticism of corpus-based measures of word similarity (as
opposed to lexicon-based measures) is that they are merely detecting
associations (co-occurrences), rather than actual semantic similarity
\cite{lund95}. To address this criticism, Lund et al. \shortcite{lund95}
evaluated their algorithm for measuring word similarity with
word pairs that were labeled \emph{similar}, \emph{associated},
or \emph{both}. These labeled pairs were originally created for
cognitive psychology experiments with human subjects \cite{chiarello90}.
Table~\ref{tab:associated} shows some examples from this
collection of 144 word pairs (48 pairs in each of the three classes).

\begin{table}[htbp]
\centering
\begin{tabular*}{0.6\linewidth}{@{\extracolsep{\fill}}ll}
\hline
Word pair & Class label \\
\hline
table:bed      & similar \\
music:art      & similar \\
hair:fur       & similar \\
house:cabin    & similar \\
cradle:baby    & associated \\
mug:beer       & associated \\
camel:hump     & associated \\
cheese:mouse   & associated \\
ale:beer       & both \\
uncle:aunt     & both \\
pepper:salt    & both \\
frown:smile    & both \\
\hline
\end{tabular*}
\caption {Examples of word pairs labeled \emph{similar}, \emph{associated},
or \emph{both}.}
\label{tab:associated}
\end{table}

Lund et al. \shortcite{lund95} did not measure the accuracy of their
algorithm on this three-class classification problem. Instead, following
standard practice in cognitive psychology, they showed that their
algorithm's similarity scores for the 144 word pairs were correlated
with the response times of human subjects in priming tests. In a typical
priming test, a human subject reads a \emph{priming} word (\emph{cradle}) and
is then asked to complete a partial word (complete \emph{bab} as \emph{baby}).
The time required to perform the task is taken to indicate the strength
of the cognitive link between the two words (\emph{cradle} and \emph{baby}).

Using ten-fold cross-validation, PairClass attains an accuracy of
77.1\% on the 144 word pairs. Since the three classes are of equal size,
guessing the majority class and random guessing both yield an
accuracy of 33.3\%. The average human score is unknown and there are no
previous results for comparison.

\section{Discussion}
\label{sec:discussion}

The four experiments are summarized in Tables \ref{tab:summary-tasks}
and \ref{tab:summary-results}. For the first two experiments, where there are previous
results, PairClass is not the best, but it performs competitively. For the second
two experiments, PairClass performs significantly above the baselines.
However, the strength of this approach is not its performance on any one task,
but the range of tasks it can handle.

\begin{table*}[htbp]
\centering
\begin{tabular*}{\textwidth}{@{\extracolsep{\fill}}lrrrrr}
\hline
Experiment & \multicolumn{2}{c}{Number of vectors} &
\multicolumn{2}{c}{Number of features} & Number of classes \\
\hline
SAT Analogies & 2,244 & ($374 \times 6$) & 44,880 & ($2,244 \times 20$) & 374 \\
TOEFL Synonyms & 320 & ($80 \times 4$) & 6,400 & ($320 \times 20$) & 2 \\
Synonyms and Antonyms & 136 & & 2,720 & ($136 \times 20$) & 2 \\
Similar, Associated, and Both & 144 & & 2,880 & ($144 \times 20$) & 3 \\
\hline
\end{tabular*}
\caption {Summary of the four tasks. See Section~\ref{sec:experiments} for explanations.}
\label{tab:summary-tasks}
\end{table*}

\begin{table*}[htbp]
\centering
\begin{tabular*}{\textwidth}{@{\extracolsep{\fill}}lccccc}
\hline
Experiment & Accuracy & Best previous & Human & Baseline & Rank \\
\hline
SAT Analogies & 52.1\% & 56.1\% & 57.0\% & 20.0\% & 2 higher out of 12 \\
TOEFL Synonyms & 76.2\% & 97.5\% & 64.5\% & 25.0\% & 8 higher out of 15 \\
Synonyms and Antonyms & 75.0\% & none & unknown & 65.4\% & none \\
Similar, Associated, and Both & 77.1\% & none & unknown & 33.3\% & none \\
\hline
\end{tabular*}
\caption {Summary of experimental results. See Section~\ref{sec:experiments} for explanations.}
\label{tab:summary-results}
\end{table*}

As far as we know, this is the first time a standard supervised
learning algorithm has been applied to any of these four problems.
The advantage of being able to cast these problems in the framework
of standard supervised learning problems is that we can now exploit
the huge literature on supervised learning. Past work on these problems
has required implicitly coding our knowledge of the nature of the
task into the structure of the algorithm. For example, the structure
of the algorithm for latent semantic analysis (LSA) implicitly contains
a theory of synonymy \cite{landauer97}. The problem with this approach
is that it can be very difficult to work out how to modify the algorithm
if it does not behave the way we want. On the other hand, with a supervised
learning algorithm, we can put our knowledge into the labeling of the
feature vectors, instead of putting it directly into the algorithm.
This makes it easier to guide the system to the desired behaviour.

With our approach to the SAT analogy questions, we are blurring
the line between supervised and unsupervised learning, since the
training set for a given SAT question consists of a single real
positive example (and a single ``virtual'' or ``simulated'' negative
example). In effect, a single example (mason:stone) becomes a
\emph{sui generis}; it constitutes a class of its own. It may be
possible to apply the machinery of supervised learning to other
problems that apparently call for unsupervised learning (for example,
clustering or measuring similarity), by using this \emph{sui generis} device.

\section{Related Work}
\label{sec:related}

One of the first papers using supervised machine learning to classify
word pairs was Rosario and Hearst's \shortcite{rosario01} paper on
classifying noun-modifier pairs in the medical domain.
For example, the noun-modifier expression \emph{brain biopsy} was
classified as \emph{Procedure}. Rosario and Hearst \shortcite{rosario01}
constructed feature vectors for each noun-modifier pair using
MeSH (Medical Subject Headings) and UMLS (Unified Medical Language System)
as lexical resources. They then trained a neural network to distinguish
13 classes of semantic relations, such as \emph{Cause}, \emph{Location},
\emph{Measure}, and \emph{Instrument}. Nastase and Szpakowicz \shortcite{nastase03}
explored a similar approach to classifying general-domain noun-modifier pairs,
using WordNet and Roget's Thesaurus as lexical resources.

Turney and Littman \shortcite{turneylittman05} used corpus-based features
for classifying noun-modifier pairs. Their
features were based on 128 hand-coded patterns. They used a nearest-neighbour
learning algorithm to classify general-domain noun-modifier pairs into
30 different classes of semantic relations. Turney \shortcite{turney06}
later addressed the same problem using 8000 automatically generated patterns.

One of the tasks in SemEval 2007 was the classification of semantic relations
between nominals \cite{girju07}. The problem is to classify semantic relations
between nouns and noun compounds in the context of a sentence. The task
attracted 14 teams who created 15 systems, all of which used supervised machine
learning with features that were lexicon-based, corpus-based, or both.

PairClass is most similar to the algorithm of Turney \shortcite{turney06},
but it differs in the following ways:

\begin{myitemize}

\item PairClass does not use a lexicon to find synonyms for the input word
pairs. One of our goals in this paper is to show that a pure corpus-based
algorithm can handle synonyms without a lexicon. This considerably simplifies
the algorithm.

\item PairClass uses a support vector machine (SVM) instead of a nearest
neighbour (NN) learning algorithm.

\item PairClass does not use the singular value decomposition (SVD) to smooth
the feature vectors. It has been our experience that SVD is not necessary with SVMs.

\item PairClass generates probability estimates, whereas Turney \shortcite{turney06}
uses a cosine measure of similarity. Probability estimates can be
readily used in further downstream processing, but cosines are less
useful.

\item The automatically generated patterns in PairClass are slightly
more general than the patterns of Turney \shortcite{turney06}.

\item The morphological processing in PairClass \cite{minnen01} is more
sophisticated than in Turney \shortcite{turney06}.

\end{myitemize}

\noindent However, we believe that the main contribution of this paper
is not PairClass itself, but the extension of supervised word pair
classification beyond the classification of noun-modifier pairs
and semantic relations between nominals, to analogies, synonyms, antonyms,
and associations. As far as we know, this has not been done before.

\section{Limitations and Future Work}
\label{sec:limitations}

The main limitation of PairClass is the need for a large corpus. Phrases
that contain a pair of words tend to be more rare than phrases that
contain either of the members of the pair, thus
a large corpus is needed to ensure that sufficient numbers of phrases are
found for each input word pair. The size of the corpus has a cost
in terms of disk space and processing time. In the future, as hardware improves,
this will become less of an issue, but there may be ways to improve
the algorithm, so that a smaller corpus is sufficient.

Another area for future work is to apply PairClass to more tasks.
WordNet includes more than a dozen semantic relations (e.g., synonyms,
hyponyms, hypernyms, meronyms, holonyms, and antonyms). PairClass
should be applicable to all of these relations. Other potential
applications include any task that involves semantic relations,
such as word sense disambiguation, information retrieval, information
extraction, and metaphor interpretation.

\section{Conclusion}
\label{sec:conclusion}

In this paper, we have described a uniform approach to analogies, synonyms,
antonyms, and associations, in which all of these phenomena are subsumed
by analogies. We view the problem of recognizing analogies as the
classification of semantic relations between words.

We believe that most of our lexical knowledge is relational, not attributional.
That is, meaning is largely about relations among words,
rather than properties of individual words, considered in isolation.
For example, consider the knowledge encoded in WordNet: much of the
knowledge in WordNet is embedded in the graph structure that connects
words.

Analogies of the form $A$:$B$::$C$:$D$ are called \emph{proportional
analogies}. These types of lower-level analogies may be contrasted with
higher-level analogies, such as the analogy between the solar system and
Rutherford's model of the atom \cite{falkenhainer89}, which are sometimes
called \emph{conceptual analogies}. We believe that the difference between
these two types is largely a matter of complexity. A higher-level analogy
is composed of many lower-level analogies. Progress
with algorithms for processing lower-level analogies will eventually
contribute to algorithms for higher-level analogies.

The idea of subsuming a broad range of semantic phenomena under analogies
has been suggested by many researchers. Minsky \shortcite{minsky86}
wrote, ``How do we ever understand anything? Almost always, I think, by
using one or another kind of analogy.'' Hofstadter \shortcite{hofstadter07}
claimed, ``all meaning comes from analogies.'' In NLP, analogical algorithms
have been applied to machine translation \cite{lepage05}, morphology \cite{lepage98},
and semantic relations \cite{turneylittman05}. Analogy provides a framework that
has the potential to unify the field of semantics. This paper is a small
step towards that goal.

\section*{Acknowledgements}

Thanks to Joel Martin and the anonymous reviewers of Coling 2008
for their helpful comments.

\bibliographystyle{coling}
\bibliography{NRC-50398}

\end{document}